\documentclass[letterpaper, 10 pt, conference]{ieeeconf}  

\IEEEoverridecommandlockouts                              

\usepackage[T1]{fontenc}
\usepackage[utf8]{inputenc}
\usepackage{authblk}
\usepackage{float}
\usepackage{cite}
\usepackage{amsmath,amssymb,amsfonts}
\usepackage{graphicx}
\usepackage{textcomp}
\usepackage{xcolor}
\usepackage{hyperref}
\usepackage{multirow}
\usepackage{hyperref}

\usepackage{algorithm}
\usepackage{algpseudocode}

\usepackage{booktabs}
\usepackage{graphicx}
\usepackage{color}
\usepackage{url}
\usepackage{soul}
\usepackage{etoolbox}
\usepackage{bbding}

\usepackage{cite}
\usepackage{balance}
\usepackage{amsmath, amssymb}
\usepackage{makecell}
\usepackage{multirow}
\usepackage{hyperref}
\usepackage{stfloats}
\usepackage{pifont}

\title{\LARGE \bf
TMR-VLA:Vision-Language-Action Model for Magnetic Motion Control of Tri-leg Silicone-based Soft Robot
}

\author{Ruijie Tang$^{*}$, Chi Kit Ng$^{*}$, Kaixuan Wu, Long Bai, Guankun Wang, \\ Yiming Huang,  Yupeng Wang, and Hongliang Ren$^{\dag}$
}

\begin{document}

\maketitle
\thispagestyle{empty}
\pagestyle{empty}

\begin{abstract}
In-vivo environments, magnetically actuated soft robots offer advantages such as wireless operation and precise control, showing promising potential for painless detection and therapeutic procedures. We developed a trileg magnetically driven soft robot (TMR) whose multi-legged design enables more flexible gaits and diverse motion patterns. For the silicone made of reconfigurable soft robots, its navigation ability can be separated into sequential motions, namely squatting, rotation, lifting a leg, walking and so on. Its motion and behavior depend on its bending shapes. To bridge motion type description and specific low-level voltage control, we introduced TMR-VLA, an end-to-end multi-modal system for a trileg magnetic soft robot capable of performing hybrid motion types, which is promising for developing a navigation ability by adapting its shape to language-constrained motion types. The TMR-VLA deploys embodied endoluminal localization ability from EndoVLA, and fuses sequential frames and natural language commands as input. Low-level voltage output is generated based on the current observation state and specific motion type description. The result shows the TMR-VLA can predict how the voltage applied to TMR will change the dynamics of a silicon-made soft robot. The TMR-VLA reached a 74\% average success rate. 
\end{abstract}

\IEEEpeerreviewmaketitle

\section{Introduction}
Miniature magnetic actuated robots, with their wireless operation, are well-suited for biomedical applications, particularly in in-vivo diagnosis and therapy~\cite{dong2022untethered}. They can reach deep inside the human body through narrow and complex pathways to perform highly precise clinical tasks. These include clearing cardiovascular blockages, delivering targeted drugs to specific lesions, and enabling minimally invasive diagnostics~\cite{dreyfus2024dexterous}. A key advantage is the compliance of soft materials, which significantly reduces the risk of collateral damage to surrounding soft tissues during internal operations. Among various actuation methods, actuation driven by an external magnetic field is widely recognized as an effective approach. This contactless mechanism enables flexible movement and precise attitude control without physical tethers—critical for navigating tortuous environments like vascular networks. Through time-varying magnetic fields generated by external coils, forces and torques can be applied to magnetic particles embedded in the robot, achieving complex deformations and locomotion. This tether-free feature is key to unlocking the full potential of soft robots in biomedical applications.

Despite promising prospects, the development of micro-scale soft robots faces a bottleneck: at micron dimensions, the robot body is too small to integrate traditional onboard power, sensors, communication modules, or control ~\cite{li2025magnetic}. Integrating such components increases the robot’s size, contradicting the goal of miniaturization. This limitation forces a separation between actuation (external magnetic fields) and perception (external observation devices). As a result, a critical gap emerges: the driver (magnetic field generator) and the robot (actuator) become physically decoupled. Unlike traditional electromechanical systems—such as motors, which use encoders for direct state feedback—external magnetic coils can only monitor their own current status. They cannot directly perceive the precise posture, deformation, or environmental interactions of the distant micro-robot. 

 With state information, controlling these systems is challenging. Soft robots are hyper-redundant, with infinite degrees of freedom. Their continuously deformable bodies exhibit complex nonlinear dynamics, defying accurate analytical modeling~\cite{haggerty2023control}. This modeling difficulty leads to a common reality: complex motions often depend on real-time human expert guidance. Experts manually adjust external magnetic fields using visual feedback, relying on experience and intuition. They become integral to the control loop~\cite{kim2022telerobotic}. A new approach is needed to add intelligence without increasing structural complexity. This would resolve the hardware limitations, eliminate reliance on human experts, and enable truly autonomous control. Therefore, we use the Vision-Language-action (VLA) framework for magnetically controlled soft robots.

The contributions of this paper are summarized as follows:
\begin{itemize}
  \item \textbf{VLA for magnetic soft robotics.} We present the first end-to-end VLA framework for magnetic soft robots, achieving hybrid primitive control using only external vision feedback.
  \item \textbf{TrilegMR-Motion dataset.} We create a \emph{TrilegMR-Motion dataset}, a vision--language--voltage dataset collected with a TMR to support learning direct voltage control from images and instructions.
  \item \textbf{Superior understanding and execution.} Compared with advanced general-purpose multimodal large language models, \emph{TMR-VLA} attains consistently higher performance in both interpreting motion instructions and successfully executing decomposed action primitives, providing a baseline for VLA in soft magnetic robots.
\end{itemize}
 
\section{Related work}

\textbf{Magnetic Configurable Miniature Robots.} Magnetically actuated soft miniature robots leverage programmed magnetization and global fields to realize rich, untethered gaits and task primitives\cite{ze2022soft}. Origami-inspired configurable designs show that field sequencing enables contraction, crawling, and steering in highly confined spaces. Programmable magnetic elastomers with multi-legs support functional integration (e.g., drug delivery, cargo transport) and robust state switching under field control~\cite{dong2022untethered}. Since global magnetic fields actuate all magnetized domains simultaneously—rendering these systems intrinsically underactuated—while embedding sensors and control electronics without compromising compliance or footprint remains difficult and magnetic feedback imposes fundamental limits on state estimation, the modeling and control of continuum magnetic soft robots confront underactuation, sparse onboard sensing, and stringent safety envelopes for medical use~\cite{li2025magnetic}. These challenges motivate us to enhance autonomy stacks that map task macros (e.g., squat/stand, leg-lift, heading correction, forward crawl) to time-varying field vectors/gradients with perceptual feedback—an approach we adopt for our silicon-based trileg robotic system that couples field actuation with voltage-induced morphology to realize sequential motion primitives.

 \textbf{Magnetically Actuated Robot Control.} The autonomy of magnetic-driven robots is predominantly concentrated in the magnetic particles or capsules, focusing on positioning and navigation~\cite{liu2024autonomous}. The application of reinforcement learning in eight-axis nonlinear magnetic drive systems achieves superior accuracy in controlling magnetic particle trajectories compared to conventional PID methods~\cite{abbasi2024autonomous}. Magnetic continuum robots can leverage existing learning-based control methods developed for continuum robotics, provided datasets are generated through dual-arm magnetic actuation~\cite{brockdorff2024hybrid}. However, for mobile robots with multi-modal locomotion (e.g., jumping~\cite{tang2024bistable}, crawling~\cite{ren2024design}, rolling~\cite{yang2024magnetic}, and shape-shifting~\cite{yang2023multimodal}), no intelligent solutions have been proposed to integrate these capabilities into task-specific frameworks systematically.

\textbf{Vision-Language-Action Model}. Vision-Language-Action (VLA) models represent a specialized category of multimodal architectures designed for embodied AI systems, integrating pre-trained vision-language foundations to jointly process visual observations~\cite{nair2022r3m}, linguistic instructions~\cite{karamcheti2023language}, and action primitives~\cite{driess2023palm}. While initial implementations~\cite{shridhar2022cliport,stone2023open} demonstrate the feasibility of embedding Vision-Language Models (VLMs) into end-to-end visuomotor policies. Subsequent advancements~\cite{huang2023embodied,li2023RoboFlamingo,zhen20243d} mitigate these issues through single-robot or simulated environment training paradigms, though such solutions sacrifice cross-platform generalizability and establish closed-loop systems resistant to efficient adaptation for novel robotic embodiments~\cite{dorka2024matters,liu2025robomamba,ding2025humanoid}. Parallel developments in visual imitation learning~\cite{wang2023demo2code,chen2024vlmimic} demonstrate alternative strategies by translating human behavioral videos into executable robotic programs. Within clinical domains, emerging research efforts have focused on adapting VLA frameworks to medical applications. In contrast to prior end-to-end low-level control, we operationalize autonomy through a \emph{library of instruction-constrained motion types} executed by a VLA policy. Each primitive (e.g., \texttt{Squat}, \texttt{Leg-lift}, \texttt{Rotation}, \texttt{Forward}) is parameterized by promptable task constraints to soft configurable robot ~\cite{kit2025endovla, wang2025trackvla}.

\begin{figure}[ht!]
\centerline{\includegraphics[width=\linewidth]{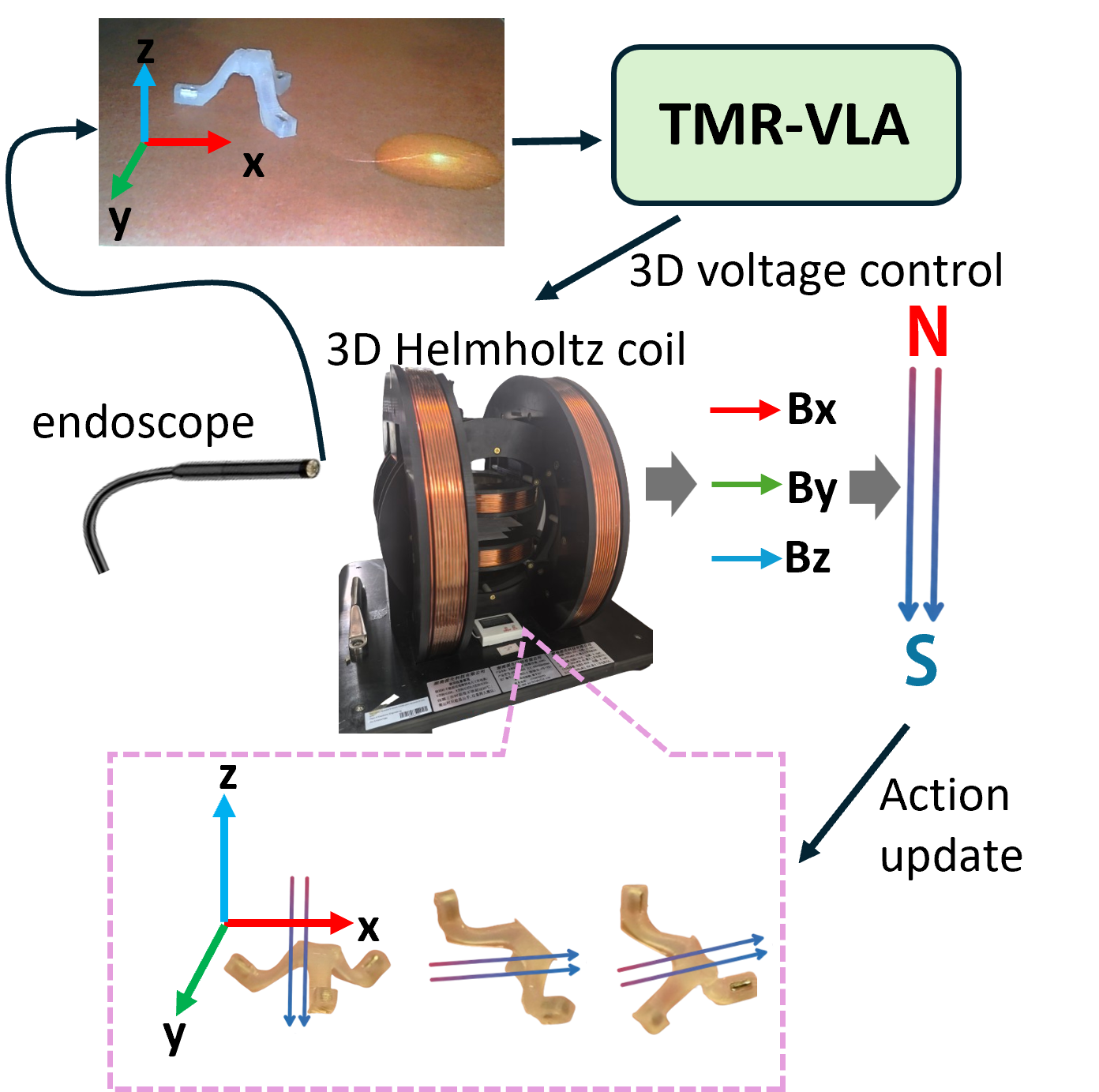}} 
\caption{The overall workflow of the VLA model for magnetic motion control of the trileg soft robot}
\label{fig1}
\end{figure}

\section{Tri-leg Magnetic Soft Robot}

\begin{figure}[ht!]
\centerline{\includegraphics[width=\linewidth]{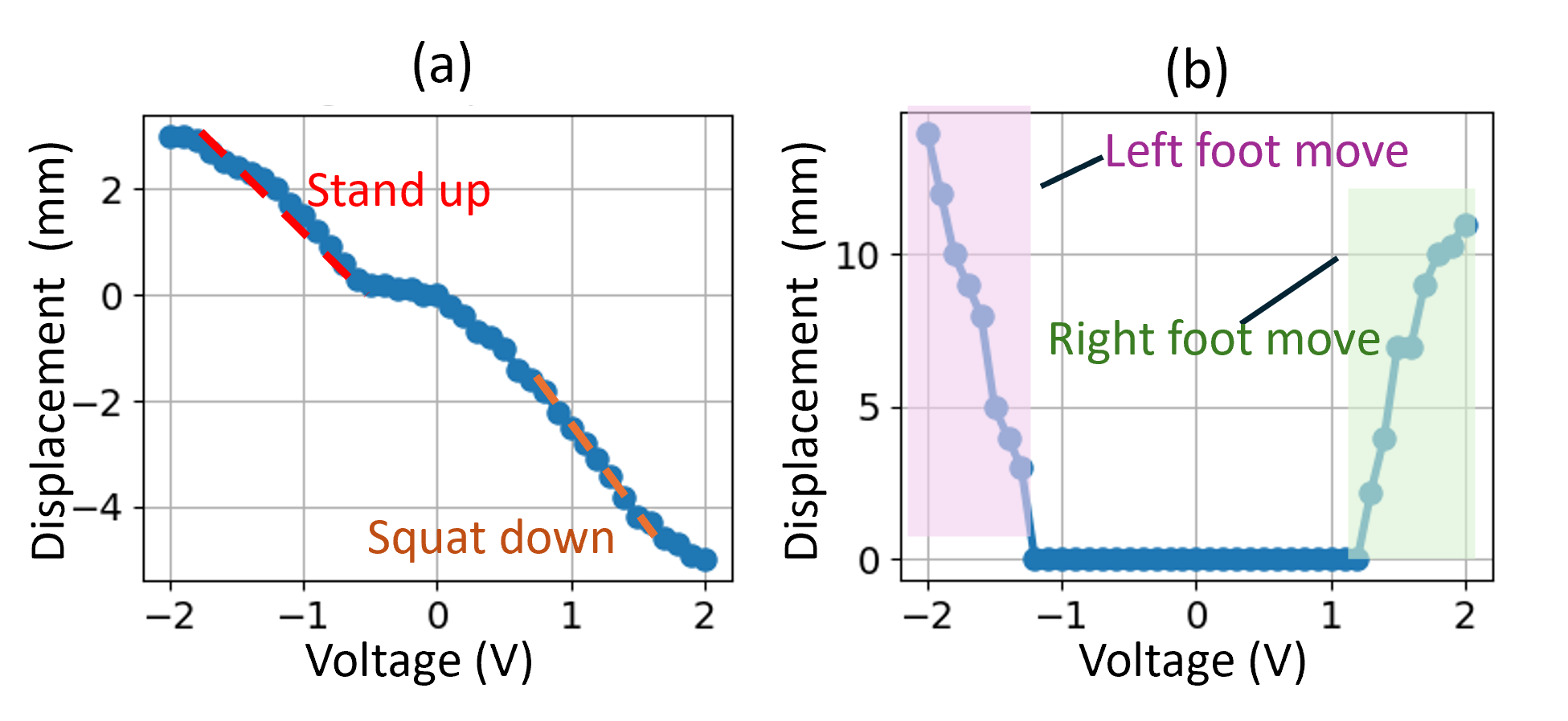}} 
\caption{Motion characterization of the Tri-leg Magnetic Soft Robot:
(a) Squatting height (z-axis) vs. voltage
(b) Stepping distance (alternating gait) vs. voltage}
\label{fig2}
\end{figure}

In~\cite{tang2024frequency}, a three-legged robot has been designed, capable of multiple motion modes such as rolling, crawling, squatting, and leg lifting. In this paper, we preliminarily discretize these motion functions to enhance generalizability for VLA training. This approach avoids relying on hard-coded API calls, which could cause the robot to deviate significantly from intended paths in unfamiliar scenarios. In principle, the robot operates within a torque generated by uniform magnetic field components along the x, y, and z axes, which counteract other possible external forces. Taking one leg as an example,
\begin{equation}
    \tau_{r}  = \mathbf{m} \times B_{r} = \mathbf{m} \times\left | B \right |\frac{\vec{r} }{\left | r \right | }, 
\end{equation}
where the $\tau_{r}$ is the torque to move the leg, $\mathbf{m}$ is the magnetic moment, $\mathbf{B}_{r}$ is the external magnetic field to move the leg. In~\cite{tang2024frequency}, to move forward in the x direction, the alternating torque ($\pm$ $\tau_{y}$) application between the legs is achieved via a sinusoidal signal ($B_{(y)} = A_{(y)} sin (2\pi ft ) $). Similarly, movement in any direction requires activating this sinusoidal signal, whose amplitude and frequency are pre-programmed. To enhance generalization capability, these parameters must be discretized into the form shown in Fig. 2. This discretization enables multimodal large language models to interpret visual scenarios and precisely execute actions across diverse situations.

\textbf{TMR motion characteristic}. The state of a soft robot is influenced by the objects it interacts with, and its own structure also determines its steady state to some extent. Therefore, in Fig~\ref{fig2}, we can observe that the squatting distance or lifting height does not increase linearly with the voltage. Only a specific segment shows a clear trend. For example, from -2 V to -0.8 V, the robot’s lifting height increases from 0 mm to 3 mm. Symmetrically, the squatting distance shows a more pronounced increase from 0 mm to 5 mm. The intermediate states, however, are not as evident. This occurs because the system surpasses a critical point in shape deformation, entering a range where bending becomes easier. For locomotion, the characterization is more complex. Fig~\ref{fig2}(b) shows measurements of two-step movements in the x+ direction. It can be observed that the robot exhibits no significant movement between -1.2 V and 1.2 V, as the driving force is below the static friction of the ground. Once the critical point is surpassed, the displacement rapidly increases to its maximum—15 mm on the left and 12 mm on the right. However, if the friction between the ground and the feet increases, the critical points described in the aforementioned curves may no longer hold. Instead, the previous action—or a sequence of past actions combined with the current state—can serve as valuable information for making judgments. The robot can leverage the pattern of these curves to interpret its own current motion, rather than relying solely on rigid thresholds measured by human experts.

\textbf{Hardware setup}. The TMR's workspace is confined to the central region of a 3D Helmholtz coil system, with dimensions of $50 mm \times 50 mm \times 30 mm$ $(L\times W \times H)$. The host computer is the terminal for receiving visual information and transmitting magnetic field signal commands. The signal generator is used to collect voltage commands and convert them into magnetic field strength.

\textbf{Data Collection}.
The commercial endoscope enters from the xz-side of the coil. It is fixed on a stand, providing a stable view. The endoscope system delivers a real-time video stream at 60 FPS. The field of view resolution can be set to either 640×480 or 1280×720. In this experiment, all data were fixed at 640×480. Due to the lack of multimodal data for soft robots without magnetic control, we created the TMRVLA-Dataset. It contains 60 episodes, totaling 15793 image-action pairs. Based on movement tasks, it is divided into three categories: 
\begin{itemize}
    \item Moving from marker A to marker B on a generic white grid background;
    \item Moving from marker A to a white lesion in a medical intestinal endoscopy setting;
    \item Moving from the current position to a yellow lesion in the same medical setting.
\end{itemize}
Data was sampled at 10 Hz, pairing endoscope images with real-time magnetic voltage values. During post-processing, redundant static voltage values and corresponding images were removed to refine the dataset. An expert manually controlled the voltage along the x, y, and z axes, using motions such as squatting, leg lifting, turning, and advancing to complete the tasks.

\section{Methodology}
\label{sec:method}

\subsection{Overview}
We formulate TMR control as end\mbox{-}to\mbox{-}end \emph{vision–language\mbox{-}to\mbox{-}voltage} prediction. At each control step $t$, the policy receives (i) a short history of overhead images $\mathcal{I}_{t-L:t}=\{I_{t-L},\ldots,I_t\}$ and (ii) a natural\mbox{-}language \emph{motion\mbox{-}type instruction} $I_m$ (e.g., \texttt{SQUAT}, \texttt{STAND\_UP}, \texttt{LIFT\_BACK\_LEG}, \texttt{ROTATE\_LEFT/RIGHT}, \texttt{FORWARD}). The policy outputs coil voltages $\Delta\mathbf{V}_t=[\Delta V_x,\Delta V_y,\Delta V_z]^\top$ that generate three orthogonal magnetic fields and induce morphology changes realizing the requested primitive. Complex tasks are executed by chaining these primitives. Our controller fuses sequential vision and language to produce low\mbox{-}level voltages tailored to the current morphology and the intended motion type constrained by the prompt.

\begin{figure}[t]
\centering
\includegraphics[width=0.86\linewidth]{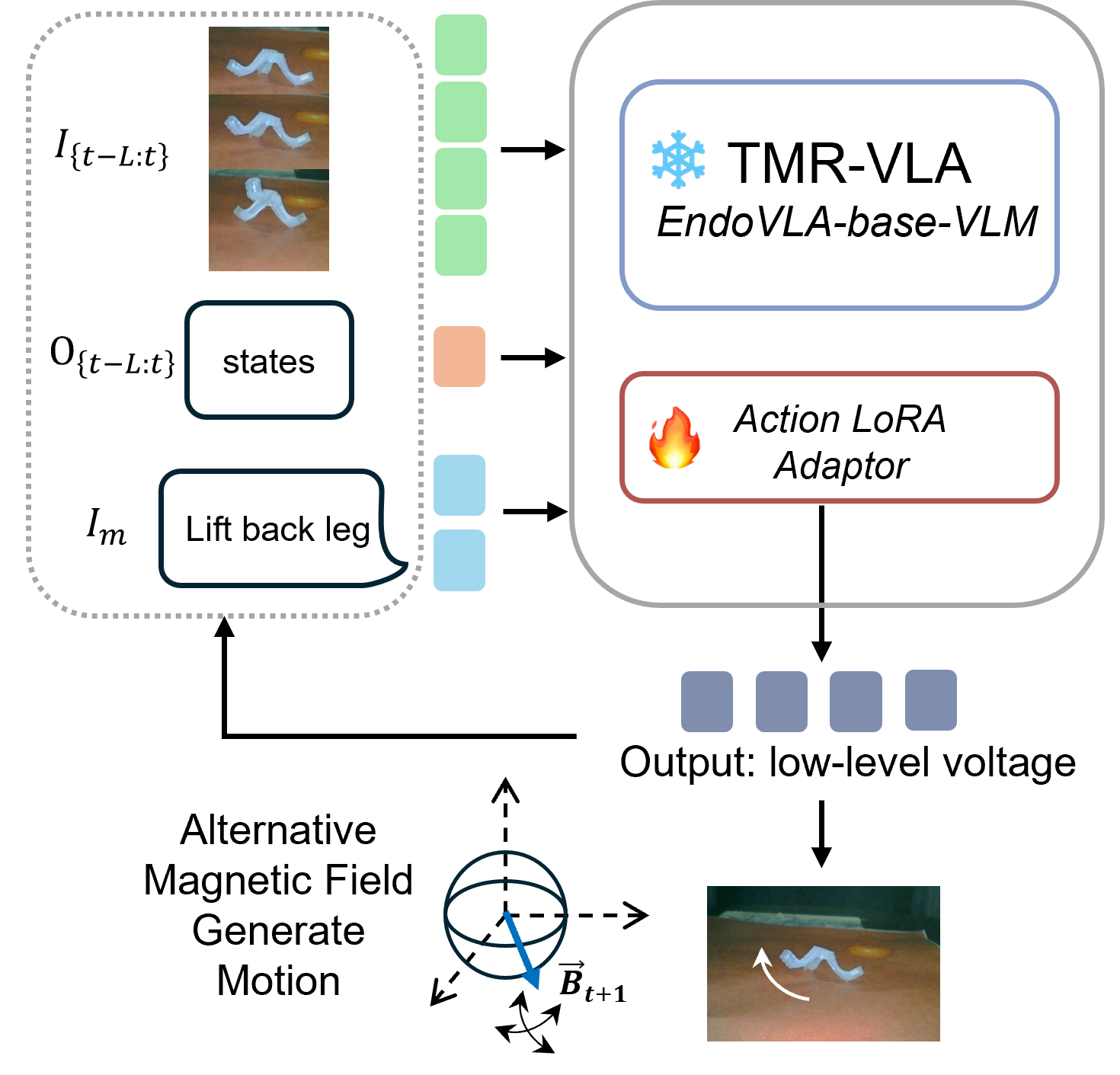}
\caption{Inference framework of the tri\mbox{-}leg magnetic robot VLA (TMR\mbox{-}VLA). The model consumes a short frame window and an instruction, then autoregressively emits quantized voltage increments that are dequantized and safety\mbox{-}projected before actuation.}
\label{fig:tmrvla}
\end{figure}

\subsection{Problem Formulation}
We cast control as a constrained partially observable MDP with observations
\begin{equation}
\label{eq:obs}
\mathbf{o}_t \,=\, \big(\mathcal{I}_{t-L:t},\ I_m,\ \mathcal{O}^{\text{aux}}_{t}\big),
\end{equation}
where $\mathcal{O}^{\text{aux}}_{t}$ denotes optional auxiliary cues (e.g., voltage state) available at training/inference time. The policy $\pi_\theta$ produces an \emph{action token sequence} $\mathbf{a}_t$ that represents a voltage increment $\Delta\mathbf{V}_t$; the applied voltage is

\begin{IEEEeqnarray}{rCl}
\mathbf{V}_t &=& \Pi_{\mathcal{C}}\!\big(\mathbf{V}_{t-1}+\Delta\mathbf{V}_t\big), \label{eq:applyV}\\
\mathcal{C} &=& \left\{\mathbf{V}\,:\ \|\mathbf{V}\|_\infty \le V_{\max},\ \|\Delta\mathbf{V}\|_\infty \le \dot V_{\max}\right\}. \nonumber
\end{IEEEeqnarray}

where $\Pi_{\mathcal{C}}$ projects onto hardware safety limits.

\paragraph*{Actuation model.}
Voltages map to magnetic fields through a calibrated coil matrix $\mathbf{K}\!\in\!\mathbb{R}^{3\times 3}$:
\begin{equation}
\label{eq:Bfield}
\mathbf{B}_t \,=\, \mathbf{K}\,\mathbf{V}_t, 
\qquad 
\boldsymbol{\tau}_t \,=\, \mathbf{m}\times \mathbf{B}_t,
\end{equation}
with effective magnetization $\mathbf{m}$ of the robot. The induced torque drives leg lifting, centroid shifting, heading correction, and forward crawling. We operate in a quasi\mbox{-}static regime (small $\Delta t$) where viscous and inertial effects are dominated by field\mbox{-}induced shape changes, and we treat the morphology state as \emph{implicitly observed} via $\mathcal{I}_{t-L:t}$.

\subsection{EndoVLA\mbox{-}Initialized Policy and Fine\mbox{-}Tuning}
We initialize the multimodal backbone from \textbf{EndoVLA}—an embodied VLA policy for endoluminal tasks~\cite{kit2025endovla}—preserving its vision encoder, text encoder, and cross\mbox{-}modal fusion blocks. We added a \emph{voltage regressor adaptor} specialized to our coil\mbox{-}driven platform:
\begin{equation}
\label{eq:map}
\big(\mathcal{I}_{t-L:t},\ I_m ,\ \mathcal{O}^{\text{aux}}_{t}\big)\ \longmapsto\ \Delta\mathbf{V}_t,
\end{equation}
and fine\mbox{-}tune by supervised behavioral cloning on $360$ paired demonstrations, minimizing a voltage regression objective with regularization (Sec.~\ref{subsec:loss}) \cite{hu2022lora}. This retains EndoVLA’s strong vision–language grounding while adapting the action mapping to magnetic soft\mbox{-}robot actuation.

\subsection{Inputs and Action Space}
\paragraph*{Vision}
A sliding window of $L$ top\mbox{-}down RGB frames captures leg poses, posture, and contact state within the coil workspace. The overhead camera substitutes for onboard sensing, consistent with miniature robots’ payload limits.

\paragraph*{Language}
A concise instruction encodes the primitive and optional constraints (e.g., “rotate left towards the marker within $5^\circ$”). Language constrains the otherwise high\mbox{-}dimensional soft\mbox{-}body behavior to a task\mbox{-}relevant subspace.

\paragraph*{Actions}
The controller outputs increments $\Delta\mathbf{V}_t=\mathbf{V}_t-\mathbf{V}_{t-1}$, bounded by hardware limits (Eq.~\eqref{eq:applyV}). Fields and torques follow Eq.~\eqref{eq:Bfield}.

\subsection{Architecture}
\paragraph*{Visual encoder and temporal aggregation.}
We use the native Qwen2.5\mbox{-}VL vision encoder—a dynamic\mbox{-}resolution ViT with \emph{Window Attention}~\cite{bai2025qwen2}. A short frame window is provided as a video clip to capture phase and contact transitions.

\paragraph*{Cross\mbox{-}modal fusion.}
Instructions are tokenized by the Qwen2.5\mbox{-}VL processor and interleaved with vision tokens using the model’s special delimiters (e.g., \texttt{<|vision\_start|>} … \texttt{<|vision\_end|>}). The concatenated stream is fed to the unmodified Qwen2.5\mbox{-}VL backbone to produce a fused representation $\mathbf{o}_t$.

\paragraph*{c) Action Adaptor}
Instead of a direct MLP regressor, we employ an \emph{action adaptor} that lets the LLM \emph{autoregressively} emit low\mbox{-}level voltage commands. Given $\mathbf{o}_t$ and $I_m$, the adaptor outputs a short token sequence encoding $\Delta\mathbf{V}_t$.

We serialize the action as
\begin{equation}
\label{eq:action_tokens}
\mathbf{a}_t=\langle\texttt{<SOS>},\ \Delta\mathbf{V}_t^{x},\ \Delta\mathbf{V}_t^{y},\ \Delta\mathbf{V}_t^{z},\ \texttt{<EOS>}\rangle,
\end{equation}
The TMR-VLA $p_\theta$ autoregressively decoding:
\begin{equation}
\label{eq:lm}
p_\theta(\mathbf{a}_t\mid\mathbf{o}_t)=\prod_{k}p_\theta\!\big(a_{t,k}\mid\mathbf{o}_t,\,a_{t<k}\big).
\end{equation}

\subsection{Training Objectives}
\label{subsec:loss}
Given demonstrations $\mathcal{D}=\{(\mathcal{I}_{t-L:t},I_m,\mathcal{O}^{\text{aux}}_t,\mathbf{V}_t)\}$, we supervise both the \emph{token} sequence and the \emph{numeric} voltages. Let $\mathbf{a}^\star_t$ be the oracle action sentence. The loss is

\begin{align}
 \label{eq:loss_total}
\mathcal{L}_{\mathrm{CE}} &= -\sum_{k}\log p_\theta\!\big(a^\star_{t,k}\mid \mathbf{o}_t, a^\star_{t,<k}\big).
\end{align}

\paragraph*{Calibration and domain randomization.}
To improve transfer across coil rigs and minor morphology changes, we \emph{abstract} robot constants (e.g., $\mathbf{K}$) into the post\mbox{-}processor and apply mild randomization to $\mathbf{K}$, and image appearance during training.

\section{Experiments}
\label{sec:experiments}

\begin{figure}[ht!]
\centerline{\includegraphics[width=\linewidth]{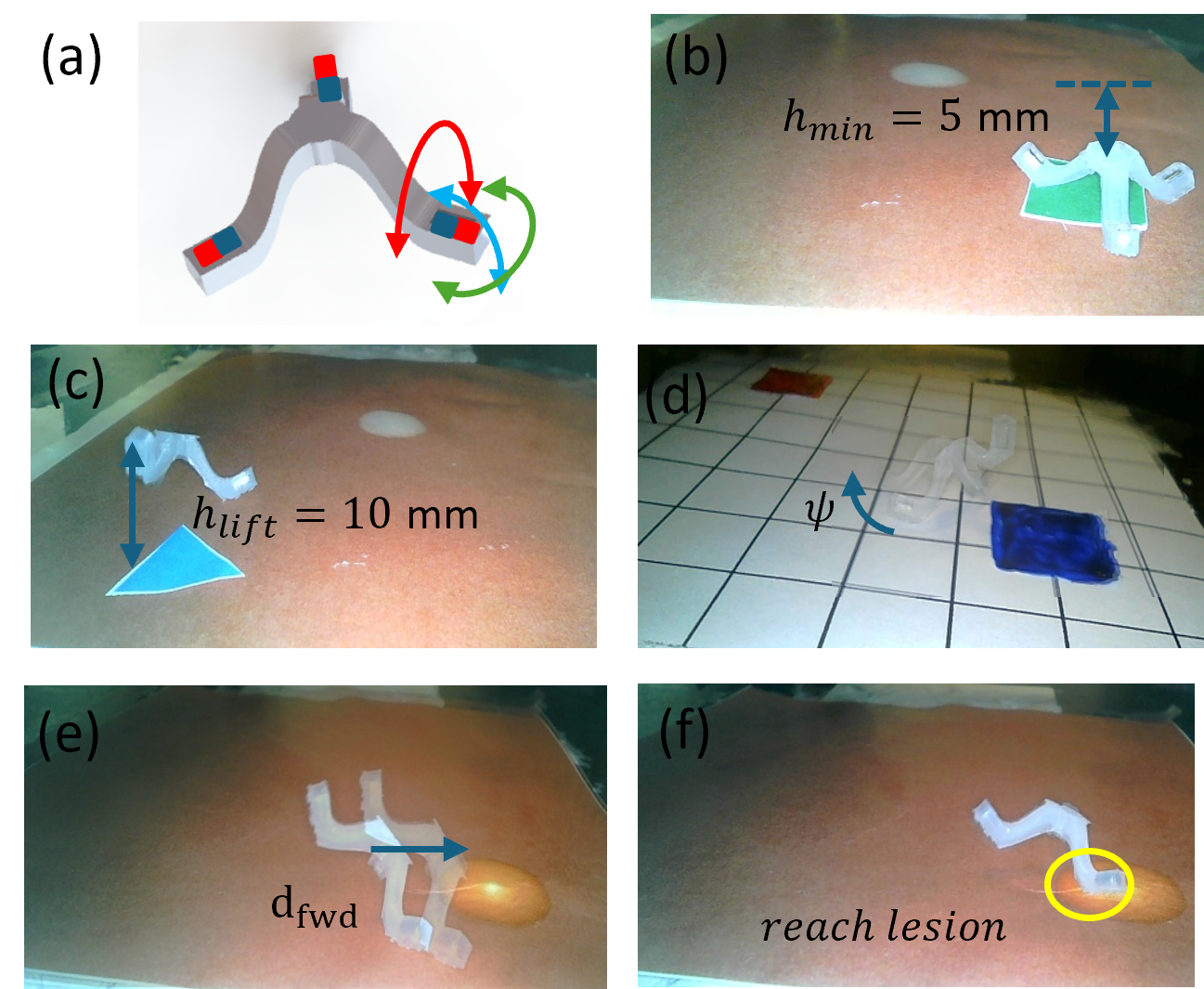}} 
\caption{(a) The illustration of the trileg robot. The magnetic field provided the torques in 3 dimensions. Experimental parameters description: (b) Maximum squat distance, (c) Leg lift height, (d) Anchor one foot, rotate body, (e) Forward distance
(f) Reach target and recover.}
\label{fig3}
\end{figure}

\subsection{Tasks and Metrics}
We evaluate five motion primitives that compose our navigation skill set:
\textbf{Squat}, \textbf{Lift-leg}, \textbf{Rotation} (left/right),
\textbf{Forward}, and \textbf{Recovery}.
Let $p_t=(x_t,y_t)$ be the 2D centroid, $\psi_t$ the heading,
and $h_\ell$ a per-leg height proxy.

\paragraph{Success criteria}
A trial succeeds if the task-specific criterion below is met within a budget $T_{\max}$ and
sustained for $T_s$ frames:
\begin{itemize}
  \item \textbf{Squat:} vertical compression $\Delta h \le -h_{\min}$ (or projected area increase $\ge A_{\min}$).
  \item \textbf{Leg-lift:} designated leg height $h_\ell \ge h_{\text{lift}}$ while the other two legs remain in contact.
  \item \textbf{Rotation:} $|\psi_T-\psi_0-\psi^\star| \le \epsilon_\psi$ and centroid drift $< d_r$.
  \item \textbf{Forward:} displacement along the commanded axis $\Delta x^{\parallel} \ge d_{\text{fwd}}$ with lateral error $|\Delta x^{\perp}| \le d_{\perp}$ and heading error $\le \epsilon_\psi$.
  \item \textbf{Recovery} designated leg returns to contact, i.e., $h_\ell \le h_{\text{drop}}$ with posture stability (centroid drift $< d_s$) and no unwanted heading change ($|\Delta\psi|\le \epsilon_\psi$).
\end{itemize}

A trial succeeds if the task-specific criterion (height/area change, heading change, or axial displacement with bounded lateral error)
is met within a budget $T_{\max}$ and sustained for $T_s$ frames.
We report per-task \emph{success rate} and the \emph{macro-average} across tasks
(Tab.~\ref{tab:motion_success}). A safety guard violation is strictly defined as exceeding the absolute voltage limit of 2.5V. Any voltage output beyond this threshold is considered a failure case. We also report the \emph{per-axis MSE} of voltage prediction. $\mathrm{MSE}_x$, $\mathrm{MSE}_y$, $\mathrm{MSE}_z$ and the \emph{overall} MSE for each motion type (Tab.~\ref{tab:mse_per_motion}).

\subsection{Hardware and Environment}
Experiments are conducted in a three-axis coil rig producing orthogonal fields
$\mathbf{B}=[B_x,B_y,B_z]^\top$ from coil voltages $\mathbf{V}=[V_x,V_y,V_z]^\top$
via a calibrated linear map $\mathbf{B}=\mathbf{K}\mathbf{V}$. The tri-leg silicone robot contains embedded magnetization; the interaction $\boldsymbol{\tau}=\mathbf{m}\times \mathbf{B}$ induces shape changes to realize the five primitives.

\subsection{Models and Training}
\paragraph{Training and Inference Implementation}
The \textbf{TMR-VLA} model was fine-tuned based on EndoVLA. Training was conducted using multiple GPUs, including four NVIDIA RTX A6000 (48 GB), and took approximately three hours to complete. We use AdamW, cosine decay, 5\% warmup. Parameter-efficient fine-tuning (LoRA/PEFT) fits a single high-memory GPU. Inference was performed on a single NVIDIA RTX 5090 (32 GB) GPU, achieving a speed of approximately 2 Hz.

\paragraph{Baselines.}
We compare against strong open models trained under the same recipe and head:
Gemma-3 (vision variant)~\cite{team2025gemma}, LLaVA-1.6~\cite{liu2023visual},
Qwen2.5-VL-3B/7B~\cite{bai2025qwen2}, and MiniCPM-V~\cite{yao2024minicpm}.
All baselines receive the same inputs (sequential frames + instruction) and
are trained to regress voltages with identical losses, schedules, and augmentations.
For backbones lacking native temporal fusion, we insert a small causal transformer over per-frame tokens.

\subsection{Evaluation Procedure}
First, we compared the action error of the baseline model and our three models against the ground truth on the test set. Then, following the previously defined success criteria, each action was executed 10 times under random initial states, and the average was taken as the success rate.

\begin{figure*}[ht!]
\centerline{\includegraphics[width=0.8\linewidth]{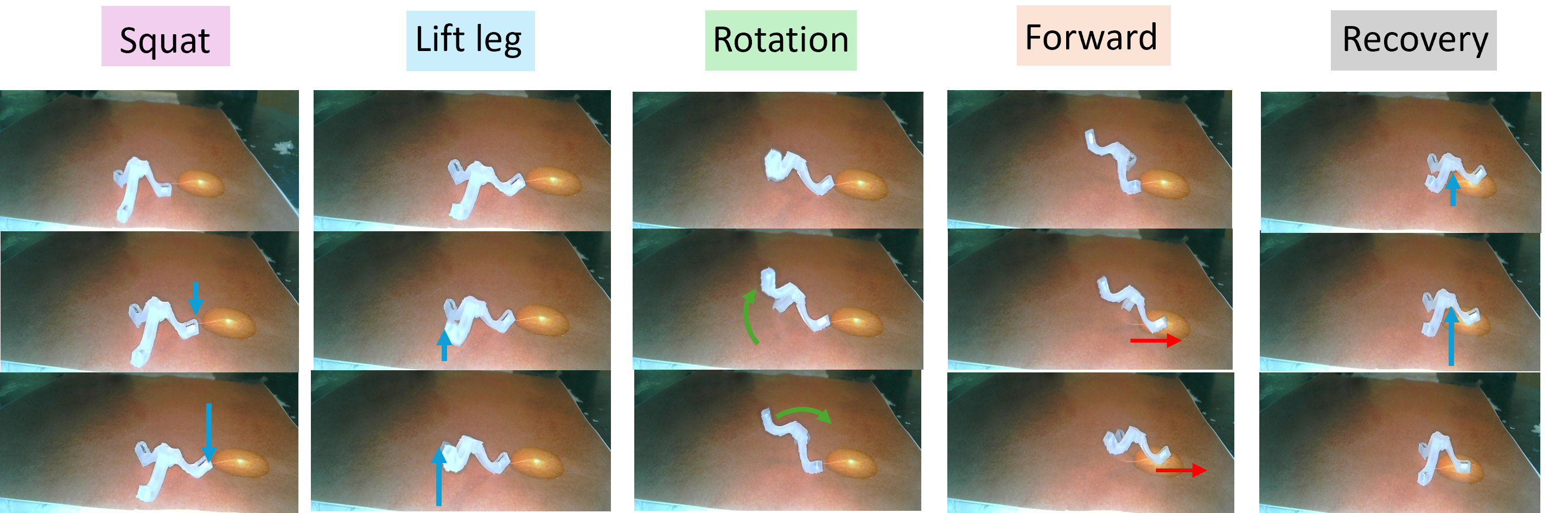}} 
\caption{Experimental results demonstrate that TMR-VLA achieves higher success rates in executing multi-step actions. The arrow indicates the magnitude and direction of the action output.}
\label{fig5}
\end{figure*}

\subsection{Prompt Design}
Table~\ref{tab:motion_success} presents the results of 
(TMR-VLA ($p_0$), TMR-VLA $(p_1)$, TMR-VLA $(p_2)$) with motion-type success rates. The system prompts for TMR-VLA $(p_0)$ and TMR-VLA $(p_1)$ are relatively concise in task description. Their common part is defined as follows: \textit{"You are a voltage controller. You can control the voltage to move the trileg robot as shown in the images. Current image is <Im0>. Past images are <Im1>, <Im2>. Current state is {s0}. Previous states are {s1} and {s2}. What is the next action?"} The key difference is that TMR-VLA $(p_0)$ includes additional historical action information: \textit{"Previous actions are {a1} and {a2}."} TMR-VLA $(p_2)$ extends TMR-VLA $(p_1)$ (which excludes historical actions) by providing a detailed task description. Its prompt is as follows: \textit{"The TMR is white with three legs, each equipped with a magnet. You control the voltage to manipulate the magnetic field, which in turn moves the robot. The robot's state represents the current strength of the magnetic field (in x, y, z directions). An action involves adjusting the magnetic field in a specific direction. For the robot to walk, it must first crouch, then lift a leg, and subsequently alternate between legs to move. You can observe the state of the robot from the images and correlate it with the numerical values."}

\subsection{Per-Motion Voltage Prediction Error (MSE)}
Table~\ref{tab:mse_per_motion} details per-axis voltage MSE ($x$, $y$, $z$) and overall MSE
for each motion type and baseline backbone.
These results complement success rates by quantifying low-level control fidelity
and highlighting axes/motions with higher actuation difficulty.

\begin{table*}[ht]
\vspace{-1.5em}
\caption{\textbf{Per-motion MSE of voltage prediction} (lower is better). Columns show $x$, $y$, $z$, and overall MSE for each motion type.}
\centering
\renewcommand{\arraystretch}{0.4}
\setlength\tabcolsep{0.35em}
\resizebox{\textwidth}{!}{
\begin{tabular}{@{}c|
cccc|cccc|cccc|cccc|cccc@{}}
\toprule
\multirow{2}{*}{\textbf{Model}} &
\multicolumn{4}{c|}{\makecell{Squat}} &

\multicolumn{4}{c|}{\makecell{Lift Leg}} &
\multicolumn{4}{c|}{\makecell{Rotation}} &
\multicolumn{4}{c|}{\makecell{Forward}} &
\multicolumn{4}{c}{\makecell{Recovery}} \\
 & \makecell{$\mathrm{MSE}_x$} & \makecell{$\mathrm{MSE}_y$} & \makecell{$\mathrm{MSE}_z$} & \makecell{Overall} 

 & \makecell{$\mathrm{MSE}_x$} & \makecell{$\mathrm{MSE}_y$} & \makecell{$\mathrm{MSE}_z$} & \makecell{Overall}
 & \makecell{$\mathrm{MSE}_x$} & \makecell{$\mathrm{MSE}_y$} & \makecell{$\mathrm{MSE}_z$} & \makecell{Overall}
 & \makecell{$\mathrm{MSE}_x$} & \makecell{$\mathrm{MSE}_y$} & \makecell{$\mathrm{MSE}_z$} & \makecell{Overall}
 & \makecell{$\mathrm{MSE}_x$} & \makecell{$\mathrm{MSE}_y$} & \makecell{$\mathrm{MSE}_z$} & \makecell{Overall} \\
\midrule
Gemma\mbox{-}3 (V) 
& \textit{0.150} & \textbf{0.000} & \textit{0.040} & \textit{0.070}

& \textit{0.015} & \textit{0.005} & \textit{0.010} & \textit{0.010}
& \textit{0.080} & \textit{0.403} & \textbf{0.000} & \textit{0.158}
& \textit{0.100} & \textit{0.100} & \textbf{0.000} & \textit{0.066}
& \textit{0.110} & \textbf{0.000} & \textit{0.050} & \textit{0.053} \\
LLaVA\mbox{-}1.6
& \textit{0.1} & \textit{0.005} & \textit{0.176} & \textit{0.093}

& \textit{0.095} & \textit{0.105} & \textbf{0.000} & \textit{0.066}
& \textit{0.093} & \textit{0.1} & \textbf{0.000} & \textit{0.064}
& \textit{0.0381} & \textit{0.108} & \textbf{0.000} & \textit{0.049}
& \textbf{0.050} & \textit{0.085} & \textit{0.050} & \textit{0.061} \\
Qwen2.5\mbox{-}VL\mbox{-}3B
& \textbf{0.004} & \textit{0.028} & \textit{0.128} & \textit{0.054}

& \textit{0.095} & \textit{0.005} & \textbf{0.000} & \textit{0.033}
& \textbf{0.078} & \textbf{0.021} & \textbf{0.000} & \textbf{0.033}
& \textbf{0.001} & \textit{0.100} & \textbf{0.000} & \textbf{0.033}
& \textbf{0.050} & \textit{0.050} & \textbf{0.000} & \textbf{0.033} \\
Qwen2.5\mbox{-}VL\mbox{-}7B
& \textit{0.009} & \textbf{0.000} & \textit{0.100} & \textit{0.036}

& \textbf{0.010} & \textit{0.005} & \textit{0.005} & \textbf{0.006}
& \textit{0.1788} & \textbf{0.021} & \textbf{0.000} & \textit{0.066}
& \textit{0.100} & \textit{0.100} & \textbf{0.000} & \textit{0.066}
& \textit{0.055} & \textbf{0.000} & \textit{0.050} & \textit{0.035} \\
MiniCPM\mbox{-}V
& \textit{0.057} & \textit{0.100} & \textit{0.143} & \textit{0.100}

& \textit{0.795} & \textit{0.105} & \textit{0.010} & \textit{0.303}
& \textbf{0.078} & \textit{0.303} & \textbf{0.000} & \textit{0.127}
& \textit{0.368} & \textit{0.313} & \textbf{0.000} & \textit{0.227}
& \textit{0.150} & \textit{0.190} & \textit{0.052} & \textit{0.134} \\

TMR-VLA ($p_0$)
& \textit{0.005} & \textbf{0.000} & \textbf{0.005} & \textbf{0.003}

& \textbf{0.010} & \textit{0.005} & \textit{0.005} & \textit{0.007}
& \textit{1.160} & \textit{1.833} & \textit{1.934} & \textit{1.644}
& \textbf{0.000} & \textbf{0.009} & \textbf{0.000} & \textbf{0.003}
& \textit{0.11} & \textit{0.007} & \textit{0.012} & \textit{0.043} \\

TMR-VLA ($p_1$)
& \textit{0.005} & \textbf{0.000} & \textbf{0.005} & \textbf{0.003}

& \textit{0.879} & \textbf{0.000} & \textit{1.158} & \textit{0.678}
& \textit{0.230} & \textit{0.209} & \textit{0.121} & \textit{0.187}
& \textit{0.008} & \textit{0.076} & \textbf{0.000} & \textit{0.028}
& \textit{0.157} & \textit{0.01} & \textit{0.095} & \textit{0.087} \\

TMR-VLA ($p_2$)
& \textbf{0.004} & \textit{0.004} & \textit{0.009} & \textit{0.006}

& \textit{0.060} & \textit{0.005} & \textit{0.045} & \textit{0.036}
& \textit{0.233} & \textit{0.2} & \textit{0.060} & \textit{0.164}
& \textbf{0.000} & \textit{0.099} & \textbf{0.000} & \textit{0.033}
& \textbf{0.050} & \textit{0.005} & \textit{0.060} & \textit{0.038} \\
\bottomrule
\end{tabular}
}
\vspace{-1em}
\label{tab:mse_per_motion}
\end{table*}

\subsection{Baseline Comparison (Success Rate)}
Table~\ref{tab:motion_success} reports motion-type success rates (\%) and the macro-average across
the five primitives for baselines and our model. We observe the relative ranking among public backbones under identical training and evaluation conditions. It should be noted that while MSE serves as a metric for evaluating action accuracy, we observed a distortion effect during experiments: when the actual action change is very small, even if the robot's output is zero, the resulting average error remains low. In contrast, the output generated by VLA control may involve random trial-and-error. Therefore, it is essential to also consider the actual experimental success rate for a comprehensive evaluation.
\setlength\tabcolsep{0.2em}
\begin{table}[ht]
\caption{\textbf{Motion type success rate (\%)}}
\centering
\renewcommand{\arraystretch}{0.4} 
\begin{tabular}{@{}c|ccccc|c@{}}
\toprule
\textbf{Model} &
\makecell{Squat} &
\makecell{Lift\\Leg} &
\makecell{Rotation} &
\makecell{Forward} &
\makecell{Recovery} &
\makecell{Mean} \\
\midrule
Gemma\mbox{-}3 (V)      & \textit{0} & \textit{0} & \textit{0} & \textit{10} & \textit{0} & \textit{2} \\
LLaVA\mbox{-}1.6        & \textit{0} & \textit{0} & \textit{20} & \textit{0} & \textit{0} & \textit{4} \\
Qwen2.5\mbox{-}VL\mbox{-}3B & \textit{10} & \textit{0} & \textit{0} & \textit{0} & \textit{0} & \textit{2} \\
Qwen2.5\mbox{-}VL\mbox{-}7B & \textit{50} & \textit{0} & \textit{10} & \textit{30} & \textit{0} & \textit{18} \\
MiniCPM\mbox{-}V        & \textit{0} & \textit{10} & \textit{0} & \textit{0} & \textit{0} & \textit{2} \\

TMR-VLA ($p_0$)        & \textbf{100} & \textit{0} & \textit{0} & \textit{60} & \textit{30} & \textit{38} \\
TMR-VLA ($p_1$)      & \textbf{100} & \textit{20} & \textit{80} & \textbf{100} & \textbf{70} & \textbf{74} \\
TMR-VLA ($p_2$)   & \textbf{100} & \textbf{50} & \textbf{100} & \textit{0} & \textit{20} & \textit{54} \\
\bottomrule
\end{tabular}
\vspace{-1em} 
\label{tab:motion_success}
\end{table}

\subsection{Robustness and Generalization}

 As shown in Table~\ref{tab:motion_success}, providing additional system descriptions does not improve the performance of the large language model in scenarios completely lacking commonsense understanding. Instead, it introduces a certain degree of interference. Comparing the results of TMR-VLA $(p_0)$ and TMR-VLA $(p_1)$, our experimental observations indicate that due to the repeated occurrence of certain directional voltage adjustments (e.g., –0.1V in the x+ direction) in the dataset, TMR-VLA tends to imitate previous actions, while TMR-VLA $(p_1)$ infers the next action based on the current state values. As a result, TMR-VLA $(p_1)$ achieves higher success rates across varying states.

\section{Discussion and Conclusion} 
\label{sec:conclusion}
Typical failures include cases where the robot returns to its initial state after completing squatting and leg lifting, without proceeding to locomotion. Additionally, when human intervention forcibly alters the model's output action, two or more modifications can cause the model to persistently follow the newly introduced action, resulting in failure. This indicates that in the absence of visual auxiliary labels such as bounding boxes or masks, the comprehension ability of large models deteriorates in non-universal scenarios. It is necessary to enable the model to genuinely understand why the robot should generate such deformations.

In conclusion, this work addresses a challenge in the development of miniature magnetically actuated soft robots: the hardware limitation that decouples external actuation from robot perception, creating a reliance on human experts for control. To bridge this gap and enable autonomous, intelligent operation without increasing structural complexity, we introduced TMR-VLA, the first end-to-end Vision-Language-Action (VLA) framework for magnetic soft robotics. Our system successfully translates high-level natural language commands and visual observations into low-level voltage control signals, predicting how applied voltages alter the state of a silicone-made tri-leg robot to execute diverse and flexible motions. Supported by the novel TrilegMR-Motion dataset, TMR-VLA demonstrated a superior ability to interpret instructions and execute hybrid motions compared to general-purpose models, achieving a 78\% average success rate. This approach provides a foundational baseline for embedding intelligence into magnetic soft robotic systems and is a significant step toward their autonomous application in complex in-vivo environments.





\newpage

\bibliography{references}
\bibliographystyle{ieeetr}

\end{document}